\pdfoutput=1

\documentclass[11pt]{article}

\usepackage[]{acl}

\usepackage{times}
\usepackage{latexsym}
\usepackage{graphicx}
\usepackage{multirow}
\usepackage{booktabs} 

\usepackage[T1]{fontenc}

\usepackage[utf8]{inputenc}

\usepackage{microtype}

\usepackage{inconsolata}

\title{SNFinLLM: Systematic and Nuanced Financial Domain Adaptation \\of Chinese Large Language Models}

\author{Shujuan Zhao$^1$, Lingfeng Qiao$^1$, Kangyang Luo$^2$, Qian-Wen Zhang$^1$, Junru Lu$^3$ and Di Yin$^1$ \\
$^1$Tecent YouTu Lab, $^2$East China Normal University, $^3$University of Warwick \\
$^1$\texttt{\{aprilzhao,leafqiao,cowenzhang,endymecyyin\}@tencent.com} \\
$^2$\texttt{52205901003@stu.ecnu.edu.cn} $^3$\texttt{junru.lu@warwick.ac.uk}}

\begin{document}
\maketitle
\begin{abstract}
Large language models (LLMs) have become powerful tools for advancing natural language processing applications in the financial industry. However, existing financial LLMs often face challenges such as hallucinations or superficial parameter training, resulting in suboptimal performance, particularly in financial computing and machine reading comprehension (MRC). To address these issues, we propose a novel large language model specifically designed for the Chinese financial domain, named SNFinLLM. SNFinLLM excels in domain-specific tasks such as answering questions, summarizing financial research reports, analyzing sentiment, and executing financial calculations.
We then perform the supervised fine-tuning~(SFT) to enhance the model's proficiency across various financial domains.
Specifically, we gather extensive financial data and create a high-quality instruction dataset composed of news articles, professional papers, and research reports of finance domain. Utilizing both domain-specific and general datasets, we proceed with continuous pre-training on an established open-source base model, resulting in SNFinLLM-base. Following this, we engage in supervised fine-tuning (SFT) to bolster the model’s capability across multiple financial tasks. Crucially, we employ a straightforward Direct Preference Optimization (DPO) method to better align the model with human preferences. Extensive experiments conducted on finance benchmarks and our evaluation dataset demonstrate that SNFinLLM markedly outperforms other state-of-the-art financial language models. For more details, check out our demo video here:\url{https://www.youtube.com/watch?v=GYT-65HZwus}.
\end{abstract}

\section{Introduction}

Recently, large language models~(LLMs) have earned profound influences on both commercial and academic spheres as the advent of ChatGPT\footnote{\url{https://openai.com/chatgpt}}. The GPT series, especially lately GPT-4 \cite{openai2024gpt4} and GPT-4o\footnote{\url{https://openai.com/index/hello-gpt-4o}}\, show remarkable performances in natural language generation~(NLG) and natural language understanding~(NLU).
These accomplishments can be attributed to these models' vast number of parameters and their training on extensive unsupervised datasets. The use of prompt-driven techniques has further refined the training process, resulting in outputs that better align with human-like responses,\cite{ouyang}.

The advancement of LLMs has not only spurred rapid development in general domains but has also facilitated the integration of LLMs into specialized domains such as medicine,\cite{zhang2023huatuogpt,sun2024medllm}, law,\cite{huang2023lawyer,cui2023chatlaw}, and finance,\cite{bloomberggpt, xuanyuan}.
These various domain tasks which can generate expected outputs by following natural prompts bring great convenience to practitioners as well as to many ordinary people. 
Here we combine the domain requirements and research result to build a financial LLM.

As we currently surveyed, there have been achievements in financial large language models.
BloombergGPT\,\cite{bloomberggpt} is a 176B English Financial LLM. Xuanyuan\,\cite{xuanyuan} is a Chinese Financial LLM with 176B parameters. Others like FinGPT\,\cite{yang2023fingpt}, DISC-FinLLM\,\cite{chen2023disc} mainly focus on implementing financial domain tasks through SFT training.
Nevertheless, the aforementioned works either merely apply parameter efficient fine-tuning methods\,\cite{devalal2018lora, ding2023peft} or lack the reinforcement learning human-alignment step\,\cite{ouyang} which may cause hallucinations and affect the performance of financial LLM.

Furthermore, while LLMs have demonstrated impressive inferential computational abilities, they still frequently encounter errors in final calculations due to inherent limitations. Inspired by ToolFormer\,\cite{schick2024toolformer}, a method of integrating tool usage within LLMs, we introduce computation tasks with calculator expressions that prompt the activation of a Python interpreter. This integration ensures accurate computations, addressing the possible wrong numerical calculations when reasoning is correct.

To this end, we introduce SNFinLLM, a specialized Language Model tailored for the Chinese financial sector, capable of addressing domain-specific inquiries, summarizing financial research, providing sentiment analysis, and performing basic financial calculations.

Our approach begins with the careful curation of a financial corpus, integrating domain-specific news, expert articles, and research reports alongside a blend of general data. Utilizing a mainstream open-source Large Language Model (LLM) as a foundation, we develop the SNFinLLM-base model through ongoing pre-training with our domain-specific dataset.

We further enhance SNFinLLM's domain expertise through Supervised Fine-Tuning (SFT). Notably, we employ Direct Preference Optimization (DPO), a straightforward but potent method referenced in \cite{rafailov2024direct}, to align SNFinLLM's reasoning capabilities more closely with human cognition.

\section{Related Work}
\subsection{LLMs}
Large Language Models (LLMs), especially ChatGPT and GPT-4\,\cite{openai2024gpt4} represent a significant advancement in natural language processing field. LLMs utilize transformers-based models and have exhibited remarkable capabilities in a range of generative tasks. InstructGPT\,\cite{ouyang} further details how to achieve natural dialogue generation by supervised fine-tuning and reinforcement learning from human feedback (RLHF). The flourishing development significantly bolsters the growth of the open-source LLMs. LLaMA\,\cite{roumeliotis2023llama} and its later version are publicly available LLMs which are trained mainly on English data. Following the continuous research, Chinese LLaMAs are proposed for excellent Chinese response abilities, like Chinese LLaMA and Alpaca\,\cite{chinese-llama-alpaca}, which empowers Chinese capacities by extending Chinese vocab and continuous pre-training on LLaMA. ChatGLM3\,\cite{zeng2022glm} is relatively early open-source LLM supporting tasks generation in both Chinese and English, which is prefix decoder-only transformers.

\subsection{Financial LLMs}

Recent advancements in financial LLMs have made significant strides. BloombergGPT\cite{bloomberggpt} stands out as the first model to leverage extensive English financial datasets, featuring a 176 billion parameter architecture and demonstrating superior performance on finance-specific tasks. The XuanYuan series\cite{xuanyuan}, comprising Chinese financial LLMs ranging from 6B to 176B parameters, has outperformed other models on the open-sourced FinanceIQ benchmark dataset. Additionally, FinGPT\cite{yang2023fingpt} employs LoRA technology to fine-tune pre-trained large language models specifically for the financial domain, effectively reducing both the number of trainable parameters and overall training costs. WeaverBird\cite{xue2023weaverbird} is another noteworthy open-source financial LLM, refined using a bilingual financial corpus in Chinese and English. Furthermore, DISC-FinLLM\cite{chen2023disc} introduces a multi-LoRA fine-tuning framework tailored to various financial tasks, enhancing model adaptability and performance.

\subsection{Computation Development in LLMs}
For strong problem-solving ability, problems with computation hope to be tackled by LLMs as well. One technique is to applying few-shot prompting\,\cite{wei2022chain, zhou2022least}. The "chain-of-thought" approach, in particular, has been shown to boost the LLMs' mathematical problems by providing them with a series of explicit intermediate steps to follow\,\cite{wei2022chain}. Despite this, LLMs are still prone to making mistakes in arithmetic computation. To address this, some studies have integrated an external calculator to handle the computations suggested by the LLMs, which has led to a modest improvement\,\cite{schick2024toolformer, chen2023disc}. The Program-Aided Language model approach\,\cite{he2023math_solving,gao2023pal} advances the concept by generating Python scripts that encapsulate the reasoning steps. These scripts are subsequently run through a Python interpreter, which guarantees accurate computations.

\begin{figure*}[ht]
\centering
\includegraphics[width=0.9\textwidth]{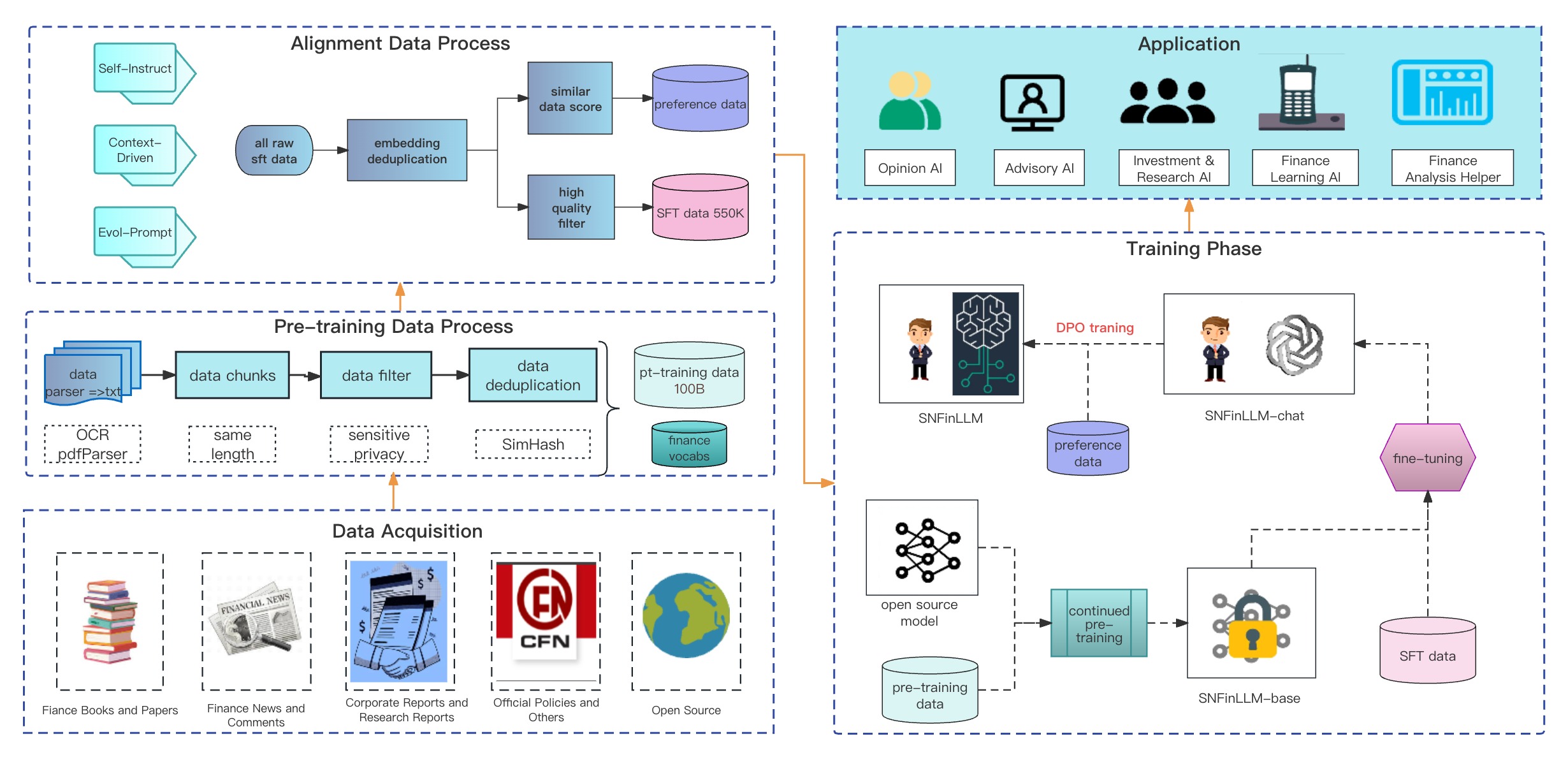} 
\caption{Framework of SNFinLLM}
\label{SNFinLLM_framework__pic:}
\end{figure*}

\section{The Approach}
We now present the overall framework of SNFinLLM (shown as Fig.~\ref{SNFinLLM_framework__pic:}), which contains five parts: data acquisition, pre-training data process, SFT data process, training phrase and application phase.
Remarkably, in the application phase, we aim to investigate the capabilities of SNFinLLM, which can serve in financial industry.
For instance, sentiment analysis is used for Opinion AI, outstanding MRC combining related documents search can be Advisory AI. Report abstracts generation and writing is for Investment \& Research AI. Question Answering (QA) tasks help finance knowledge learning and computation task can be Analysis Helper.
Thereafter, we detail the proposed SNFinLLM.

\subsection{Data Acquisition}
Data is fundamental to optimizing the performance of LLMs. In the finance domain, we encompass economics, finance, insurance, and taxation. During the data acquisition phase, our primary focus is on crawling specialized books, academic papers, news articles, financial reports, official policy documents, and relevant laws regarding insurance and taxation. Additionally, open-source datasets are vital for obtaining both domain-specific and general-purpose data. This collected data can then be utilized for domain pre-training, followed by alignment processes such as SFT and delayed policy optimization (DPO).

\subsection{Pre-training Data Process}
The pre-training data process involves four steps.  
Concretely, original data (e.g., PDF format data) is converted into plain text through tapping pdf-parser or OCR\,\cite{mithe2013OCR}  technology.
Then, long context is cut into almost same length chunks or short one concatenates to long text according to the max-length parameter.
Next, we use some rules to clean and filter the data, such as sensitive information, privacy information and other abnormal data.
Finally, SimHash method\,\cite{sadowski2007simhash} is executed for data de-duplication in the chunks level. 
In this way, we obtain a total of 25B tokens for finance unsupervised data, and together with the data carefully selected from general Chinese and English datasets, the size of all unsupervised data amounts to 100B tokens.
Note that, general Chinese and English datasets respectively have 55B and 20B tokens
being added for avoiding catastrophic forgetting\,\cite{luo2023forgetting,luo2023gradma,kirkpatrick2017overcoming}.
Also, we learned that the ratio of domain data with general data is typically 1:5\,\cite{wen2023chathome}; however, in our situation, considering the training costs and training effectiveness comprehensively, we chose a similar ratio of 1:3. Subsequent experimental results also demonstrate that a data ratio close to 1:5 is relatively reasonable. Furthermore, we employ SentencePiece tokenizer\,\cite{2018sentencepiece} to process the unsupervised financial data, resulting in the extraction of 7,689 domain-specific tokens, integrating with the existing vocabulary set to enhance overall model performance\,\cite{chinese-llama-alpaca}. 

\begin{figure*}[ht]
\centering
\includegraphics[width=0.9\textwidth]{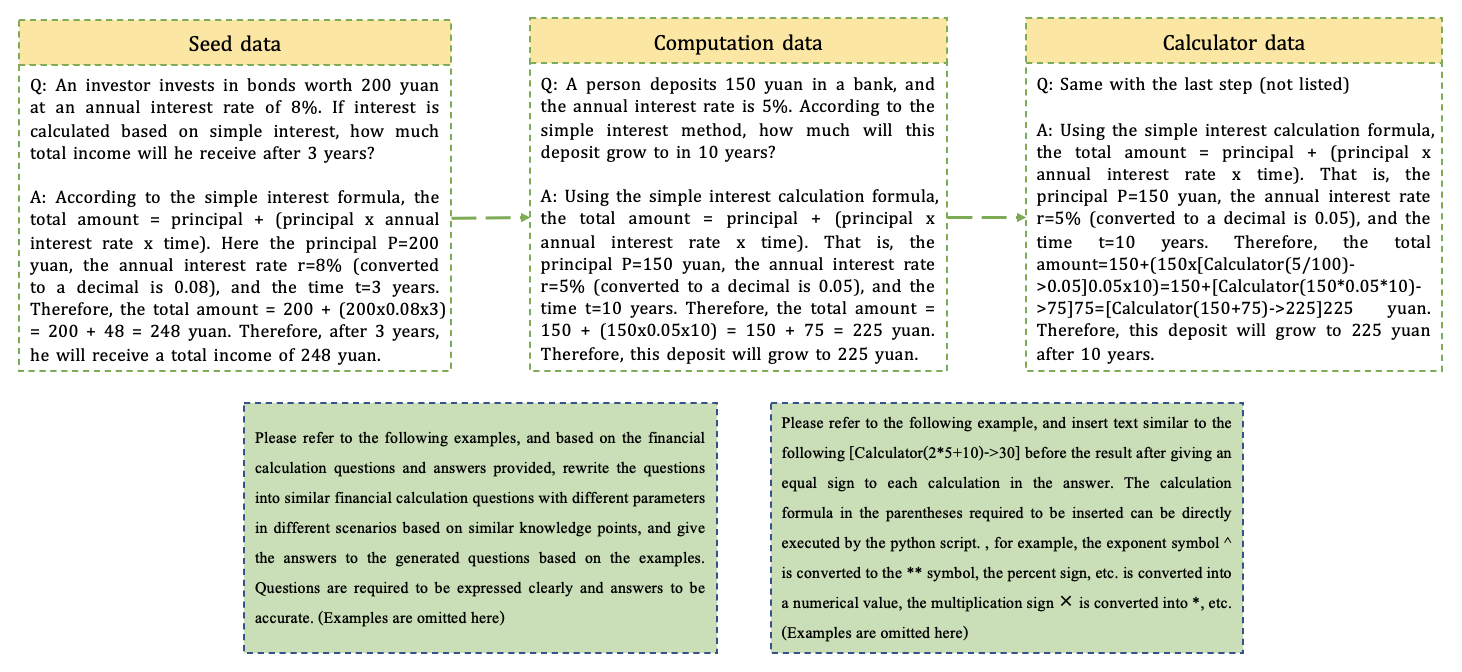} 
\caption{Steps of building computation instruction by self-instruct}
\label{com_data_build:}
\vspace*{-2ex}
\end{figure*}
\subsection{Alignment Data Process}
This phase is divided into instruction data construction and data cleaning. 
Of note, the kind of data needed is determined by the task we build. As outlined in the application phase, the SNFinLLM encompasses a variety of tasks, including QA (both multiple-choice and open-ended), financial MRC, report summarization, financial computation, sentiment analysis of comments, and relation extraction.

\subsubsection{Alignment Data Construction}
In this section, we describe the construction of financial SFT data and preference data. The latter is specifically obtained during data cleaning.

Intuitively, instruction data is constructed differently for various tasks. Building domain-specific instruction data, particularly for finance computations, can be challenging. We observe that most computation tasks are related to interest, rates, and expenses. Inspired by self-Instruct \cite{wang2022self}, we gather or generate 500 financial computation examples as seed data.
Next, we utilize GPT-4 to expand this dataset by generating similar financial data with different scenarios or parameters, along with their corresponding answers. The calculation formulas in these answers are then transformed into a format that incorporates calculator expressions, which can be executed using Python code. Fig. \ref{com_data_build:} illustrates the construction process and prompts used for GPT-4.
To ensure high-quality data, we employ regular expressions and Python interpreter checks on the complex examples, significantly reducing the need for manual rechecking.

We create Context-Driven method which leverage unsupervised data to SFT data.
For instance, we utilize analysis reports and financial news articles to create data for summarization tasks, such as generating titles or summaries. Financial concepts and their explanations are employed to formulate QAs, especially on financial terminology. Additionally, various other financial contexts are used to facilitate the construction of MRC data. 

Another crucial aspect of instruction data is to complicate the prompts to enhance the instruction following ability. We refer to this process as Evol-Prompt. Some tasks are evolved by adding constraints (e.g., output format or word number limits), while others are implemented by adding special financial worker role or what-if scenarios to the question prompts.
With the data construction phase completed, the next step is data cleaning.
\subsubsection{Data Cleaning}
In data cleaning, we first convert the supervised data into the uniform format, with instruction part, question part and answer part.
Then rules of filtering special words or symbols which are inappropriate are employed for first cleaning.

Of particular importance in the SFT data process is the task of de-duplication,\cite{dan2023educhat}. To facilitate the removal of duplicates, we embed both the instructions and questions using our self-optimized BGE model,\cite{bge_embedding}. We then employ a nearest neighbor search algorithm to identify and eliminate entries with a similarity distance below a threshold of 0.76. As a result, we achieve an 8\% reduction in the total amount of SFT data.

Furthermore, we adopt the IFD score filtering technique\,\cite{li2023ifd} to enhance the quality of our data. In this process, we choose top 70\% percent higher score data as our training dataset. Meanwhile, from low, middle and high section we filter out some data for preference data building. Ultimately, this approach yields a total of 550,000 high-quality instruction data.

In continuation, we judge answers' goodness for the similar questions got in de-duplication as preferred reply and the other one as rejected answer through GPT-4. The above filtered questions are filled in current SNFinLLM, with ground truth as preferred answer and model output as rejected one.

\subsection{Training Phase}
As displayed in Fig.~\ref{SNFinLLM_framework__pic:}, the training phase contains three steps: continued pre-training, SFT-training and DPO training. 
During the continued pre-training step, the model is fed a total of 100B tokens. Given the disparity between our data distribution and that of the original open-source base model, we conduct 3,000 warm-up steps and set the learning rate (LR) to 1e-5, which is one-tenth of the original LR. After a single epoch of training, we obtain the SNFinLLM-base model.

In the second step, we utilize the SFT data to perform full parameter fine-tuning on the SNFinLLM-base model.
In order to distinguish between common domain tasks and computation tasks generating [Calculator] expression, we tailor two system prompts. One is a prompt is: "You are one general artificial intelligence robot named ShenNong." ; the other is for calculation tasks, "You are a financial and mathematical calculation assistant. When you need to perform formula, you can call the calculator plug-in, which is a specific calculation expression can be executed by Python code. The format is [Calculator( expression)->result]." 

In this stage, we set LR to 1e-5 again, and use cosine decay learning rate scheduler with linear warmup in the first 500 steps. With one-epoch finished, we have the SNFinLLM-chat model. 

The final step is to perform DPO training in order to safely, honestly and correctly align with human's response. The parameters were configured as follows, with an LR of 1e-6 and a warmup ratio of 0.1. In this way, we obtained the final SNFinLLM.

\section{Evaluations}
We evaluate SNFinLLM on benchmarks during pre-training (PT) process, and compare SNFinLLM with SOTA finance models w.r.t. accuracy on the evaluate datasets constructed based on detailed tasks. 
For fair comparison, we ensure SNFinLLM and other models have almost equal size.

\begin{table*}[ht]
  \centering
  \caption{Evaluation results of base model and SNFinLLM series and other financial LLMs on finance benchmarks and self-evaluation datasets of instruction tasks. FEval and FIQ refer to FinEval and  FinanceIQ respectively.}
  \resizebox{1.75\columnwidth}{!}{
    \begin{tabular}{lrc|rrrrr}
    \toprule
    \multirow{2}[4]{*}{LLMs} & \multicolumn{2}{c|}{Benchmark Results} & \multicolumn{5}{c}{Self-evaluation Results} \\
\cmidrule{2-8}          & \multicolumn{1}{l}{FEval} & \multicolumn{1}{c|}{FIQ} & \multicolumn{1}{l}{qEQA} & \multicolumn{1}{l}{FinC} & \multicolumn{1}{l}{KQA} & \multicolumn{1}{l}{MRC} & \multicolumn{1}{l}{cMRC} \\
    \midrule
    Opensource-base & 59.30 & 50.32 &       &       &       &       &  \\
    \midrule
    SNFinLLM-base & \textbf{63.94} & 54.32 &       &       &       &       &  \\
    SNFinLLM-chat & 61.42 & 53.99 & 61.36 & 50.46 & \textbf{81.00}  & \textbf{95.03} & 73.20 \\
    SNFinLLM-dpo & 59.43 & 52.28 & \textbf{65.33} & 48.46 & 79.00  & 94.44 & 72.08 \\
    SNFinLLM-cal & 61.42 & 53.99 & 60.36 & \textbf{52.01} & 77.00  & 93.17 & \textbf{74.00} \\
    \midrule
    Opensource-refine & 58.22 & 48.78 & 55.24 & 40.87 & 74.00  & 87.58 & 70.10 \\
    Tongyi-Finance-14B & 62.64 & 48.57 & 51.50 & 37.77 & 73.00  & 85.09 & 72.34 \\
    XuanYuan-13B & 63.77 & \textbf{56.80} & 51.14 & 30.96 & 70.00  & 86.96 & 70.53 \\
    \bottomrule
    \end{tabular}
    }%
  \label{tab:addlabel}%
\end{table*}%

\subsection{Datasets}
During pre-training, we conduct experiments on two finance benchmark datasets which are FinEval\,\cite{zhang2023fineval} and FinanceIQ~\cite{xuanyuan}.
FinEval (Financial Evaluation) is a Chinese benchmark specifically designed for the financial domain knowledge  consisting of 4,661 multiple-choice questions covering Finance, Economy, Accounting, and Certificate\,\cite{zhang2023fineval}. 
FinanceIQ is an open-source Chinese evaluation dataset on the financial field, which collects a total of 7,173 single-choice questions covering 10 major financial categories~\cite{xuanyuan}.

In addition, as previously mentioned, we mainly focus on these supervised tasks of finance QA, practical examination, financial computation, domain MRC, summarization and finance comments sentiment analysis. For summarization task and sentiment classification showing no big difference and lack of space, we omit the two in our evaluating datasets.  Also, for fair and objective assessment, evaluators and training people are completely unconnected when preparing evaluation data.

\subsection{Benchmark Results}
\begin{figure}[t]
\centering
\includegraphics[width=0.5\textwidth]{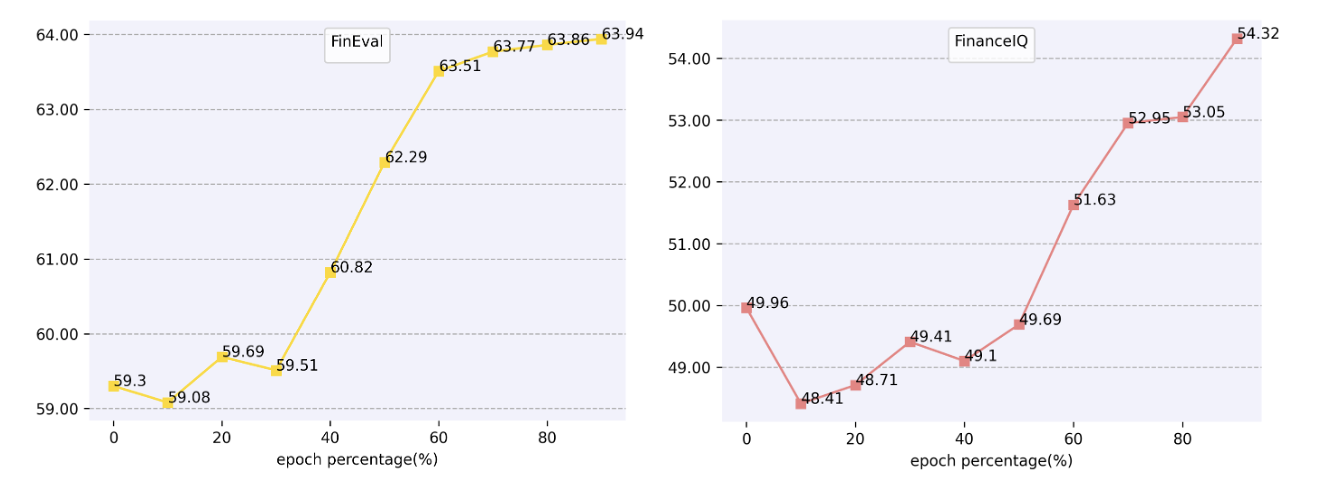} 
\caption{The Average Accuracy of SNFinLLM-base on FEval and FIQ during pre-training.}
\label{pt_eval_pic:}
\vspace*{-2ex}
\end{figure}

Here, we study the performance of different LLMs (including SNFinLLMs, the open-source base model and two finance domain chat models, which are Tongyi-Finance-14B and XuanYuan-13B) on the two benchmark datasets, as shown in Table\,\ref{tab:addlabel} Benchmark Results part. The results show that: 
SNFinLLM-base learned the finance domain knowledge in the continuous pre-training stage for FEval and FIQ gaining more than 4\% increasement compared to the original general base model. As observed in other researches, generally after SFT training, the performance value has a little decline which does not affect actual performance in instruction tasks. When evaluated on detailed instruction tasks, our SNFinLLMs exhibit markedly superior results as depicted in the Table\,\ref{tab:addlabel}.
Furthermore, we illustrate detailed evaluation results during the continued PT training in Fig. \ref{pt_eval_pic:}, demonstrating that SNFinLLM-base indeed learns about the financial domain knowledge, since the accuracy results increase as training proceeds.

\subsection{Self-evaluation Results}

We employ the self-evaluation in five internal datasets, which are qualification ExamQA (qEQA), finance computing (FinC), knowledge QA (KQA), domain MRC and complex MRC (cMRC).

From Table\,\ref{tab:addlabel} Self-evaluation Results part, we find that SNFinLLMs gain overwhelming superiority in the instruction tasks, especially the finance computing task and MRC task. 
Especially compared to Tongyi-Finance-14B and XuanYuan-13B, all results wins. Of note, in qEQA and FinC, SNFinLLM-chat achieves an almost 10\% improvement. On cMRC task, there is not great increasement, which spurs us to conduct further research for complicated context inference performance. 
Regarding the blanks in the base models' metrics, it is due to the fact that the base models lack the corresponding task-instruction capability, thus there are no metrics to fill in.

Anyway, all these success evidences our methodologies of constructing SFT data and full-parameter three stage training are beneficial to the domain LLM building. And the big gap in finance computation task also proves effectiveness in building the finance computing data and the application of calculator tool. In addition, when these SFT data trained on a larger model, the computation evaluation get to \textbf{62.54\%}, more than 12\% increased.

\subsection{Abalation Study}

From Table\ref{tab:addlabel}, we observe that the model equipped with a calculator tool (SNFinLLM-cal) achieves the best result of 52.01\% on finance computing tasks, representing an improvement of 1.55\%. A similar enhancement is seen in complex MRC tasks, which require computational steps. These experimental results demonstrate the effectiveness of integrating calculator tools for computational tasks.
The "opensource-refine" line indicates that the base model was trained on our SFT dataset without undergoing the continuous pre-training stage. The metrics highlight the necessity of domain-specific pre-training to develop a better domain-specific assistant. As shown in line 6, there is a decline of at least 4\% in performance.
Regarding the DPO training stage, a significant improvement is observed in the qEQA task. However, performance on finance computation and other tasks declines, warranting further investigation.

\section{Conclusion}
In this study, we introduce the SNFinLLM which is a Chinese financial
assistant having abilities of answering finance questions, solving finance computations, summarizing reports and other NLU tasks. The overall framework of SNFinLLM comprises data acquisition, data process, model training and application layer. The evaluation results on financial benchmark datasets as well as our built evaluating datasets showcase the effectiveness of our SFT data construction across financial tasks and the full-parameter three stage training process in the specific domain. Nonetheless, current results on qEQA and FinC are not enough for practical applications and continuing to enhance the  performance on these two tasks are required. Besides further exploration on large scale language models is another direction for the financial LLM research.

\bibliography{acl_latex}

\end{document}